\documentclass[10pt,twocolumn,letterpaper]{article}

\usepackage[pagenumbers]{cvpr} 

\usepackage[dvipsnames]{xcolor}

\definecolor{cvprblue}{rgb}{0.21,0.49,0.74}
\usepackage[pagebackref,breaklinks,colorlinks,citecolor=cvprblue]{hyperref}
\usepackage{url}
\usepackage{enumitem}
\usepackage{multirow}
\usepackage{graphicx} 
\usepackage{CJK}
\usepackage{algorithm}
\usepackage{algorithmicx}
\usepackage{algpseudocode}
\usepackage{amsmath}
\usepackage{multirow}
\usepackage{wrapfig,lipsum,booktabs}
\usepackage{tabularray}
\usepackage{xcolor}
\def\method{PETAL}

\def\paperID{3526} 
\def\confName{CVPR}
\def\confYear{2024}

\title{When Parameter-efficient Tuning Meets General-purpose \\ Vision-language Models}

\author{
\textbf{Yihang Zhai}$^{1,}$\thanks{Equal contribution.} , \textbf{Haixin Wang}$^{1,*}$, \textbf{Jianlong Chang}$^{2}$, \textbf{Xinlong Yang}$^1$, \textbf{Jinan Sun}$^{1}$, \\ \textbf{Shikun Zhang}$^1$,\textbf{Qi Tian}$^2$\\
$^1$Peking University, $^2$Huawei Cloud \& AI\\
\{zhaiyihang, wang.hx, xinlong.yang\}@stu.pku.edu.cn, \{jianlong.chang, tian.qi1\}@huawei.com,\\ 
 \{sjn, zhangsk\}@pku.edu.cn
}

\begin{document}
\maketitle
\begin{abstract}

Instruction tuning has shown promising potential for developing general-purpose AI capabilities by using large-scale pre-trained models and boosts growing research to integrate multimodal information for creative applications. However, existing works still face two main limitations: \textit{the high training costs and heavy computing resource dependence of full model fine-tuning}, and \textit{the lack of semantic information in instructions, which hinders multimodal alignment.}
Addressing these challenges, this paper proposes a novel approach to utilize \underline{\textbf{P}}arameter-\underline{\textbf{E}}fficient \underline{\textbf{T}}uning for gener\underline{\textbf{A}}l-purpose vision-\underline{\textbf{L}}anguage models, namely \textbf{PETAL}. 
\textbf{PETAL} revolutionizes the training process by requiring only 0.5\% of the total parameters, achieved through a unique mode approximation technique, which significantly reduces the training costs and reliance on heavy computing resources. Furthermore, \textbf{PETAL} enhances the semantic depth of instructions in two innovative ways: \textbf{1)} by introducing adaptive instruction mixture-of-experts(MOEs), and \textbf{2)} by fortifying the score-based linkage between parameter-efficient tuning and mutual information. Our extensive experiments across five multimodal downstream benchmarks reveal that \textbf{PETAL} not only outperforms current state-of-the-art methods in most scenarios but also surpasses full fine-tuning models in effectiveness. 
Additionally, our approach demonstrates remarkable advantages in few-shot settings, backed by comprehensive visualization analyses. Our source code is available at: \url{https://github.com/melonking32/PETAL}.
\end{abstract}    
\section{Introduction}
\label{sec:intro}

Large Language Models (LLMs) have achieved significant advances across multiple domains including language~\cite{lewis2019bart,brown2020language,touvron2023llama}, vision~\cite{rombach2022high, oquab2023dinov2} and multi-modality~\cite{Li2023BLIP2BL, dai2023instructblip}, driven by the Transformer-based architecture \cite{vaswani2017attention} and increasingly large-scale pre-training mechanism. To further enhance the model's generalization ability, more and more studies have proposed training LLMs based on instructions~\cite{wei2021finetuned, ouyang2022training}. The instruction-following manner benefits from the strong descriptive nature of language instructions, which makes models adaptable to a wide range of tasks without requiring task-specific training data and mirrors the general-purpose nature of human intelligence.

Specifically in vision-language applications, instruction tuning has been recognized as a highly effective method for boosting the performance and versatility of models. InstructBLIP~\cite{dai2023instructblip}, which quickly garners attention upon its release, uniquely fuses knowledge in pre-trained visual and language models. This technique employs human-generated instructions to precisely steer the model's actions and outputs. By utilizing the substantial semantic content within these instructions, InstructBLIP enhances the comprehension and processing of multimodal data. This approach facilitates better understanding and better interpretation of multimodal inputs, leading to improved performance in various tasks such as image captioning, visual question answering, and image generation.

\begin{figure*}[t]
  \centering
  \includegraphics[width=0.7\textwidth]{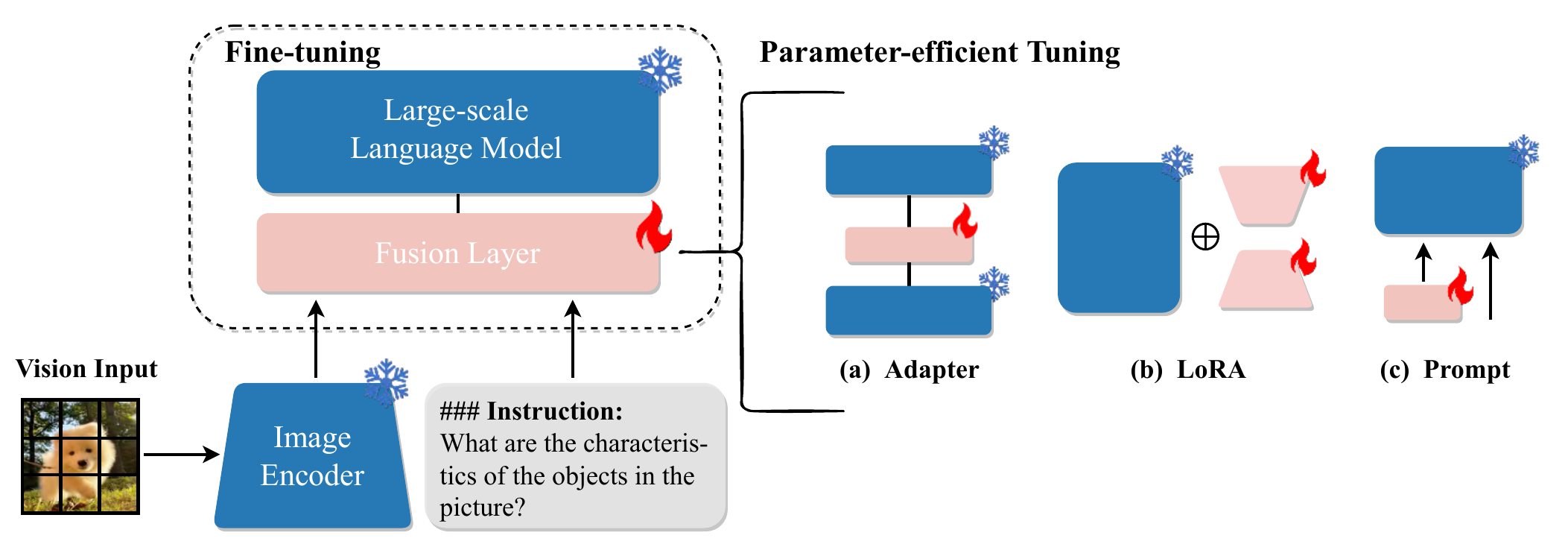}
  \caption{Overview of existing parameter-efficient tuning methods applied in multimodal tasks.}
  \label{fig:intro}
  \vspace{-0.3cm}
\end{figure*}

However, two primary challenges remain. \textbf{Firstly}, fine-tuning such large models from scratch requires massive computational resources due to their huge number of parameters (\ie, 2B for vision model, 0.2B for vision-language interface, and 3B or 7B for language model). The cost of directly fine-tuning the whole model can be prohibitively expensive. Although research has pivoted towards parameter-efficient tuning methods that significantly cut computational expenses, these strategies just include fine-tuning a limited subset of parameters or layers, employing prompt or pruning techniques without more lightweight architecture design.
\textbf{Secondly}, the richness of manually defined instruction information is limited, which weakens the fusion and alignment of multimodal information. That is to say, existing works appear in the multimodal field, such as LLaMA-Adapter~\cite{gao2023llama}, Aurora~\cite{wang2023parameter}, VL-Adapter \cite{sung2022vl}, and MAPLE \cite{khattak2022maple}, still lack of consideration in fully enhancing instruction information for aligning different modalities specifically in the context of instruct tuning.
Thus, developing new, parameter-efficient tuning techniques is vital for furthering state-of-the-art achievements on instruction tuning across diverse tasks with constrained resources.

To address this issue, we propose a novel approach to utilize \underline{\textbf{P}}arameter-\underline{\textbf{E}}fficient \underline{\textbf{T}}uning for gener\underline{\textbf{A}}l-purpose vision-\underline{\textbf{L}}anguage models, namely \textbf{PETAL}. At the core of our method are novel techniques for leveraging instruction information during parameter-efficient tuning. Specifically, PETAL first establishes an efficient fine-tuning framework based on dynamic mode approximation. It then aims to enhance the semantic information contained in instructions via an adaptive instruction MOEs module. Additionally, we introduce a mutual information loss derived from scored-based information bottleneck, which further strengthens the semantic representation of instructions. By combining parameter-efficient learning with enhanced instruction semantics, our method achieves efficient fine-tuning of large multimodal models while maintaining high performance.

Our extensive experiments, encompassing two key multimodal tasks including Image Captioning and Visual Question Answering (VQA), demonstrate the superior efficacy of our approach over previous efficient fine-tuning methods. Remarkably, our proposed PETAL registers significant performance boosts, achieving 3.5\% and 4.2\% improvements on the Flickr30K and TextCaps datasets, respectively. This is accomplished while utilizing a mere 0.5\% of the trainable parameters from the pre-trained model. This research marks a pivotal advancement in enhancing the accessibility and utility of powerful pre-trained models for a broader spectrum of applications.
The main contribution of this paper is summarized as follows:
\begin{itemize}[leftmargin=*]
    \item \textbf{Novel Efficient Tuning Framework:} Introduction of PETAL, a lightweight parameter-efficient tuning framework in general-purpose vision-language models, featuring dynamic mode approximation for efficiency.
    \item \textbf{Enhanced Instruction Semantics:} Development of an adaptive instruction MOEs module and a scored-based mutual information loss derived from a score-based information bottleneck, which collectively boost the alignment of different modalities.
    \item \textbf{Significant Performance Improvements:} Demonstrated superiority of PETAL over previous methods, with substantial performance gains on five benchmark vision-language datasets, while only using about 0.5\% of the trainable parameters from the pre-trained model.
\end{itemize}

\section{Related Work}

\subsection{Instruction Tuning}
In recent years, instruction tuning has been proven to be effective in improving the generalization ability of large language models in unseen scenarios in the field of NLP~\cite{wei2021finetuned,sanh2021multitask,ouyang2022training}. Specifically, it finetunes a pretrained large language model on a mixture of tasks phrased as instructions and can be considered as a special form of supervised fine-tuning (STF). In this paper, we focus on the generalization ability of instruction fine-tuning techniques on vision-language tasks~\cite{yang2022diffusion, yang2023improving, li2023blip, dai2023instructblip, zhu2023minigpt} and design enhanced instructions to efficiently transfer knowledge from multimodal large models to downstream tasks.

\begin{figure*}[t]
  \centering
  \includegraphics[width=0.95\textwidth]{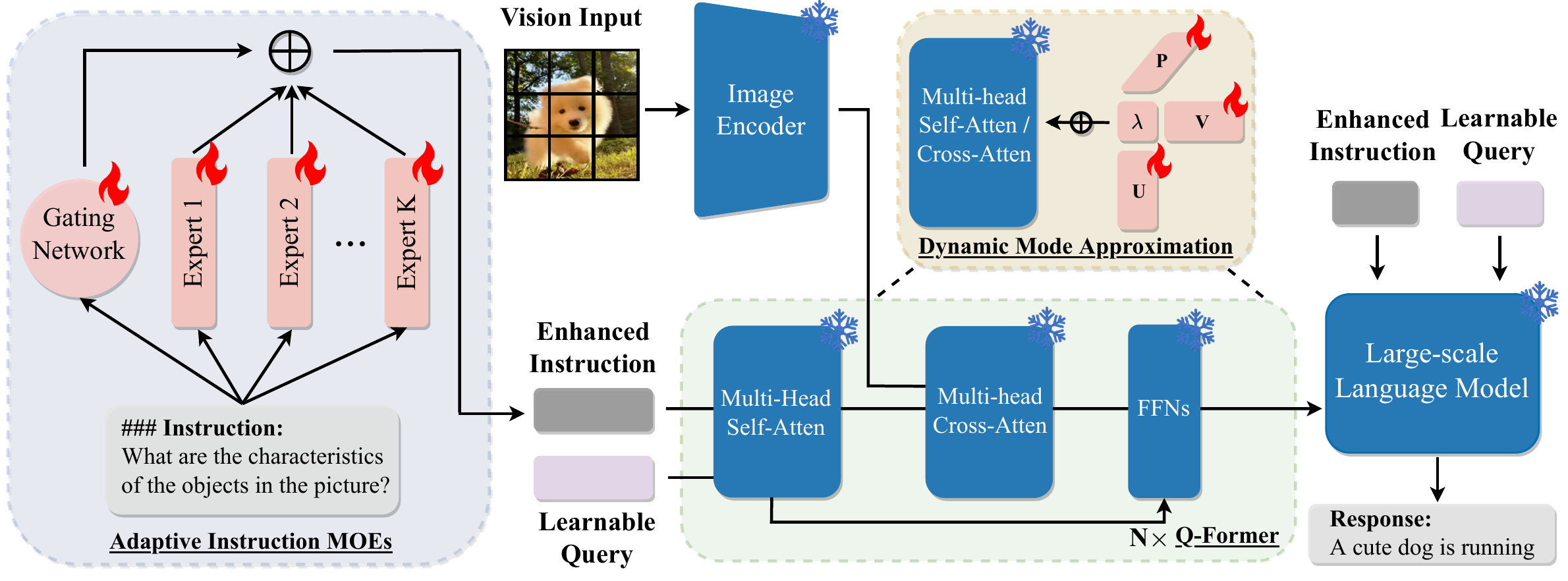}
  \caption{Demonstration of the whole framework architecture in our proposed PETAL. Our parameter-efficient tuning is only targeted at the vision-language interface, Q-Former, including both self-attention and cross-attention weight matrix. And it should be noted that \textcolor{blue}{the blue module with snow} is frozen and \textcolor{pink}{the pink module with fire} is trainable.}
  \label{fig:main}
  \vspace{-0.3cm}
\end{figure*}

\subsection{Parameter-efficient Transfer Learning}
With the development of large language models, the traditional full fine-tuning method on downstream tasks will bring huge training overhead. In this context, parameter-efficient transfer learning (PETL) has gradually become a better fine-tuning training method~\cite{ding2023parameter, yu2023visual}. Compared with updating all model parameters, PETL method only updates a small number of model parameters and transfers model knowledge to downstream tasks efficiently. There are three common categories, including prompt tuning~\citep{wang2023lion, zhou2022learning,jia2022visual,lin2023being}, adapter tuning~\cite{karimi2021compacter,chen2022adaptformer,sung2022vl,lu2023uniadapter} and LoRA-based tuning~\cite{wang2023mode,hu2021lora,jie2023fact}. In this paper, we aim to design a novel parameter-efficient tuning method for cross-modal tasks and strengthen its association with instructions.
\section{Method}




\subsection{Overall Architecture}
In this study, we adopt InstructBLIP as the foundational backbone to execute parameter-efficient tuning. Consequently, we adhere to the backbone's architecture to facilitate a streamlined design. Notably, both the visual encoder and the LLM components are kept frozen, with only the vision-language interface, termed Q-former, being designated as trainable. The comprehensive architecture of our implementation is depicted in Figure \ref{fig:main}.

\noindent\textbf{Visual Encoder.} A separate encoder ViT-g/14~\cite{Fang2022EVAET} is used as the feature extractor. It is an improved extension of the basic Vision Transformer, using 16 attention heads. It divides the input image into 14 $\times$ 14 shape-sized patches (tokens) and then encodes them using self-attention operations. 

\noindent\textbf{Q-Former.} Query-Transformer (~188M) is designed as the trainable interface to bridge the gap between a frozen LLM and a frozen image encoder \cite{dai2023instructblip}. It consists of an vision transformer submodule and a text transformer submodule, they share the same self-attention layer. Moreover, it contains a set number of learnable query embeddings. Following \cite{dai2023instructblip}, we initialize Q-Former with the pre-trained weights of \textbf{BERT}$_{base}$ \cite{devlin2018bert}, while the cross-attention layers are randomly initialized.

\noindent\textbf{LLM.} It usually has a massive number of parameters, after being trained on large-scale, high-quality text corpora. Prompts are encoded with image information as input tokens for the LLM. This activates the model to generate the desired text output. Specifically, we select FlanT5-XXL (11B) \cite{Chung2022ScalingIL} and Vicuna (7B-v1.1) \cite{vicuna2023} as the text model.

\subsection{Dynamic Mode Approximation}
\label{sec:dma}
\noindent\textbf{Traditional Low-Rank Approximation.} 
Low-rank approximation aims to find a lower-dimensional representation of a matrix that approximates the original matrix. It has applications in dimensionality reduction, model compression, and efficient computation.
Given an $m \times n$ matrix $A$, the goal is to find a rank-$r$ matrix $B$ such that $r \ll \min(m,n)$ that approximates $A$. This is done by minimizing the Frobenius norm between $A$ and $B$:

\begin{equation}
\min_{B} \left\lVert A - B \right\rVert_F
\end{equation}

Subject to the constraint that the rank of $B$ is at most $r$. where $\left\lVert \cdot \right\rVert_F$ denotes the Frobenius norm.
Find factors $W \in \mathbb{R}^{m\times r}$ and $H \in \mathbb{R}^{r\times n}$ such that $B = WH$. This is done by minimizing the objective above.
The low rank approximation $B$ provides a compressed representation of $A$ by extracting the most important latent factors.

\noindent\textbf{Dynamic Mode Approximation.} 
Motivated by \cite{wang2023parameter}, which begins by approximating the attention weights in Transformer architecture based on CP decomposition~\cite{astrid2017cp}, we improves the approximation design with a dynamic weighting scheme. Basically, three factor matrices are initialized: $U = [\mathbf{u}_1, ..., \mathbf{u}_R]$, $V = [\mathbf{v}_1, ..., \mathbf{v}_R]$ and $P = [\mathbf{p}_1, ..., \mathbf{p}_R]$. $U$ and $P$ are randomly initialized from a Gaussian distribution, while $V$ is initialized to zeros, so the initial weight change $\Delta\mathcal{W}$ is 0, and $R$ is the rank. It is worth noting that $U$ and $V$ are shared global factors used across modalities. This allows our method to consider cross-modal interactions and knowledge sharing between weight matrices. Additionally, a learnable coefficient vector $\boldsymbol{\lambda}^m \in \mathbb{R}^R$ is randomly initialized for each modality $m$ to capture discriminative features.

Besides, the concept that pre-trained knowledge encapsulated within the parameter matrix $\mathbf{W}_0$ undergoes a process of gradual distillation into the efficiently trainable parameters is widely recognized. This distillation process is critical for harnessing the inherent knowledge and capabilities of the pre-trained model, effectively transferring them into a more compact, fine-tuned parameter set. To refine and better regulate this distillation process, we propose the introduction of a dynamic learnable threshold, denoted as $\Gamma$. This threshold serves as a pivotal weighting factor for the pre-trained model, allowing for a more nuanced and controlled integration of pre-trained knowledge. During forward propagation, the weight approximation is implemented via the inverse CP decomposition process on input tensor $\mathbf{X}^m$ as:

\begin{equation}
\label{eq:dma}
\mathbf{H}^m = \Gamma \mathbf{W}_0\mathbf{X}^m + \left(\sum_{r=1}^R \lambda_r^m (\mathbf{u}_r \circ \mathbf{v}_r \circ \mathbf{p}_r)\right) \mathbf{X}^m
\end{equation}

This decomposes the weight tensor into the product of three factor matrices and a coefficient vector per modality. And the dynamic nature of $\Gamma$ is particularly noteworthy as it enables dynamic modulation of the influence exerted by the pre-trained model throughout the training process. 

\subsection{Adaptive Instruction MOEs}
In current approaches, instructions typically originate from human-authored datasets~\cite{wang2022self}, which are often constrained in scope, variety, and originality. These constraints present notable challenges and restrict the overall adaptability of the model. Conversely, tailoring instructions to directly correspond with the images, such as identifying entities within the photographs, can significantly enhance the integration of linguistic and visual elements, aligning more closely with natural human perception. Additionally, image characteristics span multiple dimensions, each emphasizing different facets. Relying solely on a singular instruction can lead to a skewed emphasis on specific image features, overlooking the comprehensive perspective.

A common way to address the problem is the use of deep sparse Mixture-of-Experts (MOEs) \cite{riquelme2021scaling,mustafa2022multimodal,shen2023scaling}. By selectively
activating only the relevant portions, MOEs can enhance the model representation ability without a corresponding rise in computational expense. However, directly applying MOEs to instruction tuning is difficult to achieve in two folds: (1) Instruction is inherently general-purpose, making it challenging to define a large set of universally applicable instructions across various tasks. (2) Generations of instructions heavily rely on human expertise, leading to high costs even when utilizing existing pre-trained models.

To extract information from multiple perspectives, we propose a novel method called adaptive instruction MOEs for instruction tuning, which is presented in Figure \ref{fig:main}.
Firstly, we set multiple instructions for each image, each with a different focus on the image, and stack them line by line in a text paragraph. The features extracted are diverse, and by merging these features, we implement a mechanism similar to voting, aiding in capturing a more comprehensive understanding of the image. Secondly, we feed the instruction $x$ to $K$ experts with an adaptive gating network. Each of these experts are composed of a single non-linear transformation layer as follows: 
\begin{equation}
    y_k = \gamma_k \odot x + \beta_k
\end{equation}
where $\gamma_k$ and $\beta_k$ is the transformation factor of the $k-$th expert. Finally, the gating network $\mathcal{G}$ is architecturally designed as a neural network, functioning to adaptively allocate weights to each expert based on the input features. The gating network computes the attention weights as follows:
\begin{equation}
g_k = \frac{\exp \left(\mathcal{G}(x, y_k)\right)}{\sum_{j=1}^K \exp(\mathcal{G}(x, y_j) )}
\end{equation}
By merging these outputs, the final enhanced instruction representation is obtained as $I_{in} = \sum_{k=1}^K g_k y_k $. 

\begin{table*}[t]
\caption{We select two captioning datasets, Flickr30K and TextCaps, for performance comparison on the famous image captioning benchmark. We report CIDEr score, ROUGE-1 F1 score, ROUGE-1 recall, ROUGE-L F1 score, and ROUGE-1 recall for both of them, where \textcolor{red}{$\uparrow$} and \textcolor{green}{$\downarrow$} respectively indicates how much our method has improved or declined compared to the best parameter-efficient baseline.}
\centering
\scriptsize
\label{tab:caption}
\vspace{-0.3cm}
\setlength{\tabcolsep}{3mm}{
\begin{tabular}{ll|ccccc|ccccc}
\toprule[1pt]
 &  & \multicolumn{5}{c|}{\textbf{Flickr30K}} & \multicolumn{5}{c}{\textbf{TextCaps}} \\ \cline{3-12}
 &  & {\color[HTML]{202122} CIDEr} & \multicolumn{2}{c}{{\color[HTML]{333333} ROGUE-1}} & \multicolumn{2}{c|}{ROGUE-L} & {\color[HTML]{202122} CIDEr} & \multicolumn{2}{c}{{\color[HTML]{333333} ROGUE-1}} & \multicolumn{2}{c}{ROGUE-L} \\ \cline{3-12}
\multirow{-3}{*}{Method} & \multirow{-3}{*}{Tunable} &  & F1 & Recall & F1 & Recall & &  F1 & Recall & F1 & Recall \\ \midrule
Fine-tune InstructBLIP (FlanT5$_{\text{xxl}}$) & 188 M & \textbf{63.5} & 34.9 & 33.4 & 31.6 & 30.4 & 46.6 & 28.2 & 26.3 & 24.7 & 23.0 \\ \midrule
Head (FlanT5$_{\text{xxl}}$) & 11.8 M & 60.8 & 34.5 & 31.9 & 30.7 & 29.2 & 43.6 & 28.9 & \textbf{28.8} & 24.6 & 24.5 \\
MAPLE (FlanT5$_{\text{xxl}}$) & 2.9M & 59.4 & 34.2 & 31.8 & 30.8 & 28.1 & 43.7 & 27.8 & 26.5 & 24.3 & 22.4 \\
LLaMA-Adapter (FlanT5$_{\text{xxl}}$,R=128) & 4.8 M & 60.5 & 33.7 & 31.2 & 30.5 & 28.3 & 44.5 & 28.3 & 23.9 & 24.5 & 23.1 \\
LoRA (FlanT5$_{\text{xxl}}$,R=64) & 5.0 M & 59.8 & 34.0 & 31.5 & 30.8 & 28.6 & 45.4 & 28.0 & 25.5 & 24.7 & 22.5 \\
PETAL (FlanT5$_{\text{xxl}}$,R=64) & \textbf{1 M} & 63.3 & \textbf{35.2} & \textbf{32.9} & \textbf{35.2} & \textbf{32.9} & \textbf{46.7} & \textbf{29.1} & 27.8 & \textbf{29.1} & \textbf{28.5} \\ \midrule
 &  & {2.5\textcolor{red}{$\uparrow$}} & {0.7\textcolor{red}{$\uparrow$}} & {1.0\textcolor{red}{$\uparrow$}} & {4.2\textcolor{red}{$\uparrow$}} & {3.7\textcolor{red}{$\uparrow$}} & {1.3\textcolor{red}{$\uparrow$}} & {0.2\textcolor{red}{$\uparrow$}} & {1.0\textcolor{green}{$\downarrow$}} & {4.4\textcolor{red}{$\uparrow$}} & {4.0\textcolor{red}{$\uparrow$}} \\ \midrule
Fine-tune InstructBLIP (Vicuna-7B) & 188 M & \textbf{65.8} & 35.5 & 34.3 & \textbf{36.0} & 34.3 & \textbf{48.9} & 30.8 & \textbf{31.5} & 30.9 & \textbf{31.5} \\
LLaVA (Vicuna-7B) & 7B & \multicolumn{1}{c}{64.6} & \multicolumn{1}{c}{\textbf{35.9}} & \multicolumn{1}{c}{33.7} & \multicolumn{1}{c}{34.2} & \multicolumn{1}{c|}{34.0} & \multicolumn{1}{c}{48.5} & \multicolumn{1}{c}{30.9} & \multicolumn{1}{c}{27.9} & \multicolumn{1}{c}{28.7} & \multicolumn{1}{c}{-} \\ \midrule
Head (Vicuna-7B) & 11.8 M & \multicolumn{1}{c}{61.9} & \multicolumn{1}{c}{35.1} & \multicolumn{1}{c}{32.9} & \multicolumn{1}{c}{33.8} & \multicolumn{1}{c|}{33.7} & \multicolumn{1}{c}{48.3} & \multicolumn{1}{c}{30.3} & \multicolumn{1}{c}{30.6} & \multicolumn{1}{c}{30.3} & \multicolumn{1}{c}{30.6} \\
MAPLE (Vicuna-7B) & 2.9M & \multicolumn{1}{c}{61.3} & \multicolumn{1}{c}{35.6} & \multicolumn{1}{c}{33.0} & \multicolumn{1}{c}{35.1} & \multicolumn{1}{c|}{32.4} & \multicolumn{1}{c}{48.1} & \multicolumn{1}{c}{30.5} & \multicolumn{1}{c}{30.8} & \multicolumn{1}{c}{30.7} & \multicolumn{1}{c}{30.1} \\
LLaMA-Adapter (Vicuna-7B,R=128) & 4.8 M & \multicolumn{1}{c}{62.4} & \multicolumn{1}{c}{36.6} & \multicolumn{1}{c}{33.2} & \multicolumn{1}{c}{33.4} & \multicolumn{1}{c|}{32.9} & \multicolumn{1}{c}{48.3} & \multicolumn{1}{c}{30.2} & \multicolumn{1}{c}{30.0} & \multicolumn{1}{c}{30.4} & \multicolumn{1}{c}{30.2} \\
LoRA (Vicuna-7B,R=64) & 5.0 M & 62.5 & 35.0 & 33.8 & 35.0 & 33.9 & 48.5 & 30.8 & 31.2 & 30.8 & 31.2 \\
PETAL (Vicuna-7B,R=64) & \textbf{1 M} & 63.6 & 35.7 & \textbf{34.3} & 36.2 & \textbf{34.3} & \textbf{48.8} & \textbf{31.1} & 30.5 & \textbf{32.0} & 31.3 \\ \midrule
 &  & {1.1\textcolor{red}{$\uparrow$}} & {0.4\textcolor{green}{$\downarrow$}} & {0.5\textcolor{red}{$\uparrow$}} & {0.6\textcolor{red}{$\uparrow$}} & {0.4\textcolor{red}{$\uparrow$}} & {0.3\textcolor{red}{$\uparrow$}} & {0.3\textcolor{red}{$\uparrow$}} & {0.7\textcolor{green}{$\downarrow$}} & {1.2\textcolor{red}{$\uparrow$}} & {0.1\textcolor{red}{$\uparrow$}} \\ \bottomrule[1pt]
\end{tabular}}
\vspace{-0.3cm}
\end{table*}

\subsection{Scored-based Information Bottleneck.} 
To enhance the integration of instruction and vision information during parameter-efficient tuning, we propose the score-based information bottleneck module. In the context of Q-former's efforts to merge multimodal data, the key lies in directing the learning process more effectively. We achieve this by strategically amalgamating task-relevant details from instructions, ensuring the retention of their semantic connections. This is done through the calculation of normalized attention weights, denoted as $\alpha_i$, which leads to the formation of a unified representation, symbolized as $\hat{H}$. Additionally, to effectively combine instruction-specific features, represented as ${{z}_i}$, while maintaining their inherent semantic links, we assign a correlation weight $\alpha_i$ to each feature. This method fosters a more efficient and cohesive blending of instruction and visual data, leveraging the strengths of score-based information bottleneck in parameter-efficient tuning scenarios as follows:

\begin{equation}
\alpha_i = \text{SoftMax}(H_i^T H_j)
\end{equation}

The weighted features $\hat{{H}} = \sum_i \alpha_i {H}_i$ are then used to calculate an information bottleneck loss. The goal of the information bottleneck is to find a representation $Z$ of input $X$ that preserves as much information as possible about a target variable $Y$, while losing as much information as possible about the input itself.
It aims to maximize the mutual information $I(Z;Y)$ between the representation $Z$ and target $Y$, while minimizing the mutual information $I(Z;X)$ between the representation and the input:

\begin{equation}
\max \text{MI} = \max I(Z; Y) - \eta I(Z; X)
\end{equation}

Where $\eta$ is a hyper-parameter that controls the trade-off between preserving information about $Y$ versus losing information about $X$. The mutual information can be expressed as:

\begin{equation}
I(Z; Y) = \sum_{z,y} p(z,y) \log \frac{p(z|y)}{p(z)}
\end{equation}

\begin{equation}
I(Z; X) = \sum_{z,x} p(z,x) \log \frac{p(z|x)}{p(z)}
\end{equation}

Where $p(z|y)$ is the conditional probability of representation $z$ given target $y$, $p(z)$ is the marginal probability of $t$, and similar for $p(z|x)$.
Finally, the loss calculated together with the labels can be formed as:

\begin{equation}
\label{eq:IB}
\mathcal{L}_{IB} = \text{MI} (\hat{\mathbf{z}}; y)
\end{equation}

This loss plays an important role in enhancing the semantics of instructions to better guide the parameter-efficient tuning process. Optimizing the overall loss (including the cross-entropy loss in InstructBLIP) enables our method to efficiently fine-tune large multimodal models with enhanced performances. 
\section{Experiments}

\subsection{Experimental Setup}

\smallskip\noindent\textbf{Datasets.} This research utilizes a diverse array of benchmark datasets, summarized below:
(1)
\textbf{Flickr30K} \cite{Young2014FromID} comprises 31K images sourced from Flickr, each accompanied by five reference captions provided by human annotators, enriching the dataset with varied descriptive perspectives.
(2)
\textbf{TextCaps} \cite{Sidorov2020TextCapsAD} features 28K images derived from the open images database, and each image is complemented by five human-generated captions, emphasizing textual descriptions within the visual content.
(3)
\textbf{OKVQA} \cite{Marino2019OKVQAAV} presents a unique set of visual questions that demand external knowledge for accurate responses. The dataset is partitioned into 9K training and 5K test samples.
(4)
\textbf{A-OKVQA} \cite{Schwenk2022AOKVQAAB}, an advanced iteration of OKVQA. It includes 17K training, 1K validation, and 6K test questions, each designed to push the boundaries of a model's reasoning capabilities.
(5)
\textbf{TextVQA} \cite{Singh2019TowardsVM} tests the model's ability to interpret visual text as part of the question-answering process. It contains a comprehensive collection of 35K training questions and an additional 5K for testing.

\smallskip\noindent\textbf{Baselines.} A wide range of state-of-the-art methods are utilized to make the comparisons, including two full fine-tune methods (\ie InstructBLIP~\cite{dai2023instructblip}, and LLaVA~\cite{liu2023improvedllava}.) and four parameter-efficient tuning methods (\ie Head Tuning~\cite{zaken2021bitfit}, MAPLE~\cite{khattak2022maple}, LLaMA-Adapter~\cite{gao2023llama} and LoRA~\cite{hu2021lora}.) For fair comparison, we re-implement all the baselines by extracting the core code from their parameter-efficient tuning section under our instruction tuning settings.

\noindent\textbf{Implementation Details.} We froze the parameters across the entire backbone architecture, opting to apply parameter-efficient tuning exclusively to the Q-Former enhanced by approximation techniques. Our LLMs of choice are FlanT5$_{\text{xxl}}$ and Vicuna-7B.
The experimental procedures were conducted using the LAVIS library~\cite{li2022lavis}. All experiments are instruction-tuned with five epochs on 8$\times$ Nvidia A100 (80G) GPUs.
For the training details, we set a batch size of 32. Besides, we use the AdamW optimizer with $\beta1 = 0.9$, $\beta2 = 0.999$, and a weight decay of 0.02. We further refine the training by incorporating a linear warm-up for the learning rate over the first 1,000 steps, 
increasing from $1e\text{-}6$ to $2e\text{-}5$, followed by a cosine decay.

\subsection{Main Results} 
\subsubsection{Results on Captioning Task}
Image captioning aims to capture the shallow and deep information of the image as comprehensively as possible.
In our experiments, we conduct training on both the Flickr30K and TextCaps captioning datasets, and evaluate our model on the test/validation sets. 
To ensure robust results, we utilize metrics such as CIDEr Score \cite{Vedantam2014CIDErCI}, ROGUE \cite{Lin2004ROUGEAP}, etc. 
Appendix provides a detailed introduction to the evaluation metrics we use.

As shown in Table \ref{tab:caption}, \method{} utilizes only 0.5\% of the parameters (about 1M introduced in Appendix) and has significantly outperformed other efficient tuning methods on almost all metrics across two datasets. 
Especially when using FlanT5$_{\text{xxl}}$ framework, we exceeded the previous best method by 18\% in ROGUE score, and even even surpasses full-parameter fine-tuning methods.
When using the Vicuna-7B framework, our advantages are still evident.
These results demonstrate the strong capability of our approach in effectively extracting image information and fusing it with textual information. 

\noindent\textbf{Qualitative Evaluation} 
In order to further investigate the effectiveness of our approach in real-world scenarios 
, we select various types of images for qualitative analysis. As depicted in Figure \ref{fig:case1} of the Appendix, \method{} effectively captures key entity information and human actions in the image, whereas the outputs of other methods, while containing image information, lack sufficient features and miss some critical information. This also indicates that our 
method can capture more comprehensive information from images across multiple dimensions.
In Figure \ref{fig:case4} of the Appendix, it can be observed that other methods provide shallow descriptions focusing only on the most prominent parts of the image. They lack analysis of dynamic information within the image and the relationships between different entities. In contrast, \method{} 
accurately identifies hidden dynamic information and potential relationships between entities. The results align with human intuition.

\subsubsection{Results on VQA Task}
To further explore the potential of PETAL in other types of multimodal tasks, we evaluate it on A-OKVQA, OKVQA, and TextVQA datasets commonly employed in VQA tasks.  The evaluation results are presented in Table \ref{tab:vqa}. VQA tasks typically necessitate external knowledge, making them comparatively challenging. However, our proposed \method{} consistently outperforms other parameter-efficient tuning methods in the majority of scenarios, providing additional evidence for the robust generalization capabilities of \method{} in the context of visual question-answering tasks. It should noted that our PETAL even surpass the Fine-tune InstructBLIP in some scenarios.

Results from the two experimental sets presented above comprehensively illustrate the effectiveness and generalization capability of PETAL in various multimodal tasks. These findings provide a robust and viable solution for the efficient fine-tuning of PETAL in a broader range of general-purpose multimodal tasks.

\begin{table}[t]
\caption{Performance comparison on VQA benchmarks. We conduct experiments on three datasets: A-OKVQA, OKVQA, and TextVQA, using accuracy as the evaluation metric. \textcolor{red}{$\uparrow$} and \textcolor{green}{$\downarrow$} respectively indicates our improvement compared to the best baseline.}
\vspace{-0.3cm}
\centering
\scriptsize
\label{tab:vqa}
\setlength{\tabcolsep}{3mm}{
\begin{tabular}{ll|ccc}
\toprule[1pt]
 & \multicolumn{1}{c|}{} &  &  &  \\
\multirow{-2}{*}{Method} & \multicolumn{1}{c|}{\multirow{-2}{*}{Tunable}} & \multirow{-2}{*}{A-OKVQA} & \multirow{-2}{*}{TextVQA} & \multirow{-2}{*}{OKVQA} \\ \hline
\multicolumn{5}{l}{\textbf{Based on FlanT5$_{\text{xxl}}$}}  \\ \hline
Fine-tune & 188 M & {\color[HTML]{262626} \textbf{56.7}} & {\color[HTML]{262626} \textbf{24.1}} & \textbf{55.2} \\
Head & 11.8 M & 54.3 & 21.0 & 52.1 \\
MAPLE & \multicolumn{1}{l|}{2.9M} & 54.1 & 21.2 & 52.4 \\
LLaMA-Adapter & 4.8 M & {\color[HTML]{262626} 53.2} & {\color[HTML]{262626} 20.9} & 52.8 \\
LoRA  & 5.0 M & {\color[HTML]{262626} 54.5} & {\color[HTML]{262626} 21.4} & 53.4 \\
PETAL & 1.0 M & 55.8 & {\color[HTML]{262626} 21.5} & 53.6 \\ \hline
 &  & {\textcolor{red}{$\uparrow$}1.3} & {\textcolor{red}{$\uparrow$}0.1} & {\textcolor{red}{$\uparrow$}0.2} \\ \hline
 \multicolumn{5}{l}{\textbf{Based on Vicuna-7B}}  \\ \hline
 Fine-tune & \multicolumn{1}{l|}{188 M} & {\color[HTML]{262626} 63.6} & {\textbf{62.4}} & 27.7 \\
 LLaVA & \multicolumn{1}{l|}{7B} & 52.2 & 52.7 & 21.3 \\
Head & \multicolumn{1}{l|}{11.8 M} & 63.2 & 60.1 & 25.8 \\
MAPLE & \multicolumn{1}{l|}{2.9M} & {\color[HTML]{262626} 62.8} & {\color[HTML]{262626} 60.3} & 24.5 \\
LLaMA-Adapter & \multicolumn{1}{l|}{4.8 M} & {\color[HTML]{262626} 63.0} & {\color[HTML]{262626} 60.5} & 26.4 \\
LoRA  & \multicolumn{1}{l|}{5.0 M} & {\color[HTML]{262626} 63.5} & {\color[HTML]{262626} 60.8} & 26.3 \\
PETAL & \multicolumn{1}{l|}{1.0 M} & \textbf{63.8} & {61.8} & \textbf{27.7} \\ \hline
 & \multicolumn{1}{l|}{} & {\textcolor{red}{$\uparrow$} 0.3} & {\textcolor{red}{$\uparrow$} 1.0} & {\textcolor{red}{$\uparrow$} 1.3} \\ \bottomrule[1pt]
\end{tabular}}
\vspace{-0.5cm}
\end{table}

\begin{table*}[t]
\caption{Results of few-shot instruction tuning. We have two configurations: we extract 50/150 data items from the training set for training. We conducted tests on AOKVQA, OKVQA, FLickr30K, and TextCaps. For AOKVQA and OKVQA, we calculate the accuracy, while for Flickr30K and TextCaps, we compute the CIDEr Score and ROGUE-1 F1 Score.}
\label{tab:fewshot}
\centering
\scriptsize
\vspace{-0.3cm}
\setlength{\tabcolsep}{4mm}{
\begin{tabular}{lccccccccc}
\toprule[1pt]
\multicolumn{1}{l}{} &\multicolumn{1}{c|}{} & {\color[HTML]{202122} \textbf{A-OKVQA}} & \multicolumn{1}{c}{\textbf{OKVQA}} & \multicolumn{1}{c|}{} & \multicolumn{2}{c}{\textbf{Flickr30K}} & \multicolumn{2}{c}{\textbf{TextCaps}} & \multicolumn{1}{c}{}\\ \cmidrule{3-10}
\multicolumn{1}{l}{\multirow{-2}{*}{Method}} & \multicolumn{1}{c|}{\multirow{-2}{*}{Parameter}} &Accuracy & Accuracy & \multicolumn{1}{c|}{Avg} & CIDEr & ROGUE-1 F1 & CIDEr & ROGUE-1 F1 & \multicolumn{1}{c}{Avg} \\ \midrule
\textbf{50-shot} &  &  &  &  &  & & &  \\ \midrule
\multicolumn{1}{l}{Fine-tuning} & \multicolumn{1}{c|}{188 M} & 53.2 & 52.0 & \multicolumn{1}{c|}{52.6} & \textbf{53.5} & 33.1 & 42.8 & 27.1 & 39.1 \\
\multicolumn{1}{l}{LoRA} & \multicolumn{1}{c|}{5 M} & 53.1 & 52.1 & \multicolumn{1}{c|}{52.6} & 52.7 & 33.1 & 43.0 & 27.1 & 38.8 \\
\multicolumn{1}{l}{Ours} & \multicolumn{1}{c|}{1 M} & \textbf{53.3} & \textbf{52.6} & \multicolumn{1}{c|}{\textbf{53.0}} & 52.9 & \textbf{33.2} & \textbf{43.1} & \textbf{27.4} & \textbf{39.2} \\ \midrule
\textbf{150-shot} &  &  &  &  &  & & &  \\ \midrule
\multicolumn{1}{l}{Fine-tuning} & \multicolumn{1}{c|}{188 M} & 53.0 & 52.2 & \multicolumn{1}{c|}{52.6} & \textbf{53.5} & 33.2 & 42.7 & 27.1 & 39.1 \\
\multicolumn{1}{l}{LoRA} & \multicolumn{1}{c|}{5 M} & 53.1 & 52.1 & \multicolumn{1}{c|}{52.6} & 52.7 & 33.1 & 42.9 & 27.0 & 38.9 \\
\multicolumn{1}{l}{Ours} & \multicolumn{1}{c|}{1 M} & \textbf{53.1} & \textbf{52.6} & \multicolumn{1}{c|}{\textbf{52.9}} & 53.1 & \textbf{33.2} & \textbf{43.2} & \textbf{27.4} & \textbf{39.2} \\ \bottomrule[1pt]
\end{tabular}}
\vspace{-0.3cm}
\end{table*}

\begin{table*}[t]
\caption{Results of ablation studies that remove the important components. \textcolor{red}{$\uparrow$} and \textcolor{green}{$\downarrow$} respectively indicates how much the variant has improved or declined compared to our PETAL. DMA stands for Dynamic Mode Approximation, AIM represents Adaptive Instruction MOEs, and SIB is Score-based Information Bottleneck loss.}
\centering
\label{tab:ablation}
\scriptsize
\vspace{-0.3cm}
\setlength{\tabcolsep}{5.8mm}{
\begin{tabular}{lccc|c|ccc|c}
\toprule[1pt]
 & & & &\textbf{A-OKVQA} & \multicolumn{3}{c|}{\textbf{Flickr30K}} &  \\ \cline{5-8}
 & & & &Accuracy  & CIDEr & \multicolumn{2}{c|}{ROGUE-L} &  \\ \cline{5-8}
\multirow{-3}{*}{Method} & \multirow{-3}{*}{DMA} & \multirow{-3}{*}{AIM} & \multirow{-3}{*}{SIB}& \multirow{-2}{*}{} & \multirow{-2}{*}{} & {\color[HTML]{333333} F1} & {\color[HTML]{333333} Recall} & \multirow{-3}{*}{Avg.} \\ \midrule
\textbf{PETAL} V1& $\checkmark$ & $\times$ & $\checkmark$& 55.2 & 61.1 & 34.5 & 32.3 & 45.8 (\textcolor{green}{$\downarrow$}1.1) \\
\textbf{PETAL} V2 & $\checkmark$ & $\checkmark$ & $\times$ & 55.6 & 62.6 & 34.8 & 33.1 & 46.5 (\textcolor{green}{$\downarrow$}0.4) \\
\textbf{PETAL} V3 & $\checkmark$ & $\times$& $\times$ & 54.8 & 60.7 & 31.0 & 28.9 & 43.9 (\textcolor{green}{$\downarrow$}3.0 ) \\ 
\textbf{PETAL} V4 & $\times$ & $\checkmark$ & $\checkmark$ & 55.4 & 61.2 & 33.8 & 31.9 & 45.6 (\textcolor{green}{$\downarrow$}1.3) \\ \midrule
\multicolumn{4}{l|}{\textbf{PETAL} w. random instruction}& 55.1 & 60.8 & 32.0 & 30.9 & 44.8 (\textcolor{green}{$\downarrow$}2.2) \\ \midrule
\textbf{PETAL}& $\checkmark$ & $\checkmark$ & $\checkmark$ & \textbf{55.8} & \textbf{63.4} & \textbf{35.1} & \textbf{33.4} & \textbf{46.9} \\ \bottomrule[1pt]
\end{tabular}}
\vspace{-0.3cm}
\end{table*}

\subsubsection{Results on Few-shot Task}
Few-shot learning and efficient fine-tuning share common objectives, namely, the efficient adaptation of pre-trained models to specific downstream tasks. The potential synergy achieved by combining these two methodologies holds promise for further enhancing model efficiency. In this context, our primary objective is to investigate the generalization capabilities of the PETAL.

We achieve this by conducting rigorous testing of our approach within a few-shot learning scenario.
In our experiments, we adhere to a principled approach by initializing the model with the parameters derived from InstructBLIP. This deliberate choice ensures the equitable comparison of different methods, mitigating uncertainties associated with varying initial conditions and enabling a more precise evaluation of method performance.
Then we conduct experiments on four distinct datasets: AOKVQA, OKVQA, Flickr30K, and TextCaps. To emulate the conditions of few-shot learning, we select two settings characterized by a limited number of training samples, consisting of 50 samples and 150 samples, respectively.

Based on the experimental results (refer to Table~\ref{tab:fewshot}), it is evident that PETAL consistently outperforms other efficient fine-tuning techniques across all datasets. Moreover, its performance closely approximates that of traditional fine-tuning methods even better. These findings underscore the exceptional generalization capabilities of PETAL within the context of few-shot learning.
The significance of this generalization capability cannot be overstated, as it implies that PETAL can be widely applied across a diverse spectrum of downstream tasks without necessitating extensive retraining. This attribute serves to enhance the practical utility and efficiency of the model.
These experimental results, along with their accompanying analysis, provide further substantiation of the practicality and generalization capabilities of the PETAL, positioning it as a promising solution for efficient few-shot learning scenarios.

\begin{figure}[t]
    \centering
    \includegraphics[width=0.49\textwidth]{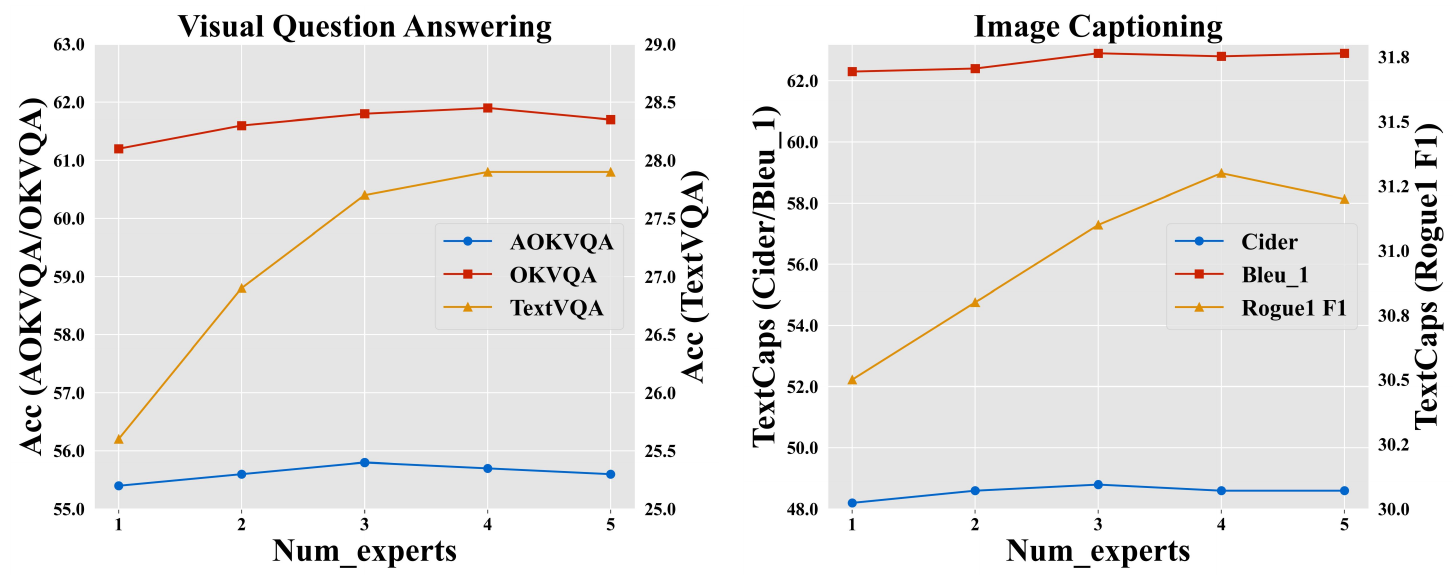}
    \vspace{-0.3cm}
    \caption{Results of different number of experts.}
    \label{fig:experts} 
    \vspace{-0.5cm}
\end{figure}

\subsection{Sensitivity Analysis}
In this section, we delve into the impact of varying the number of experts, as depicted in Figure \ref{fig:experts}. Our analysis reveals that in the realm of Visual Question Answering, there is a marked uptick in accuracy when the expert count is escalated from one to three, after which the gains plateau, indicating a threshold beyond which additional experts do not equate to substantial enhancements. Conversely, in the field of Image Captioning, we note a nuanced response: CIDEr scores ascend with a growing roster of experts, whereas Bleu-1 scores initially rise before tapering off, and the ROGUE-L F1 scores peak with three experts, hinting at an optimal number for maintaining sentence structure fidelity.
These trends indicate that while the inclusion of more experts generally leads to better performance, there is a complex interplay between the number of experts and the specific nature of the task. Thus, the optimal number of experts is set to 3 as default.

\begin{figure*}[t]
    \centering
    \includegraphics[width=0.7\textwidth]{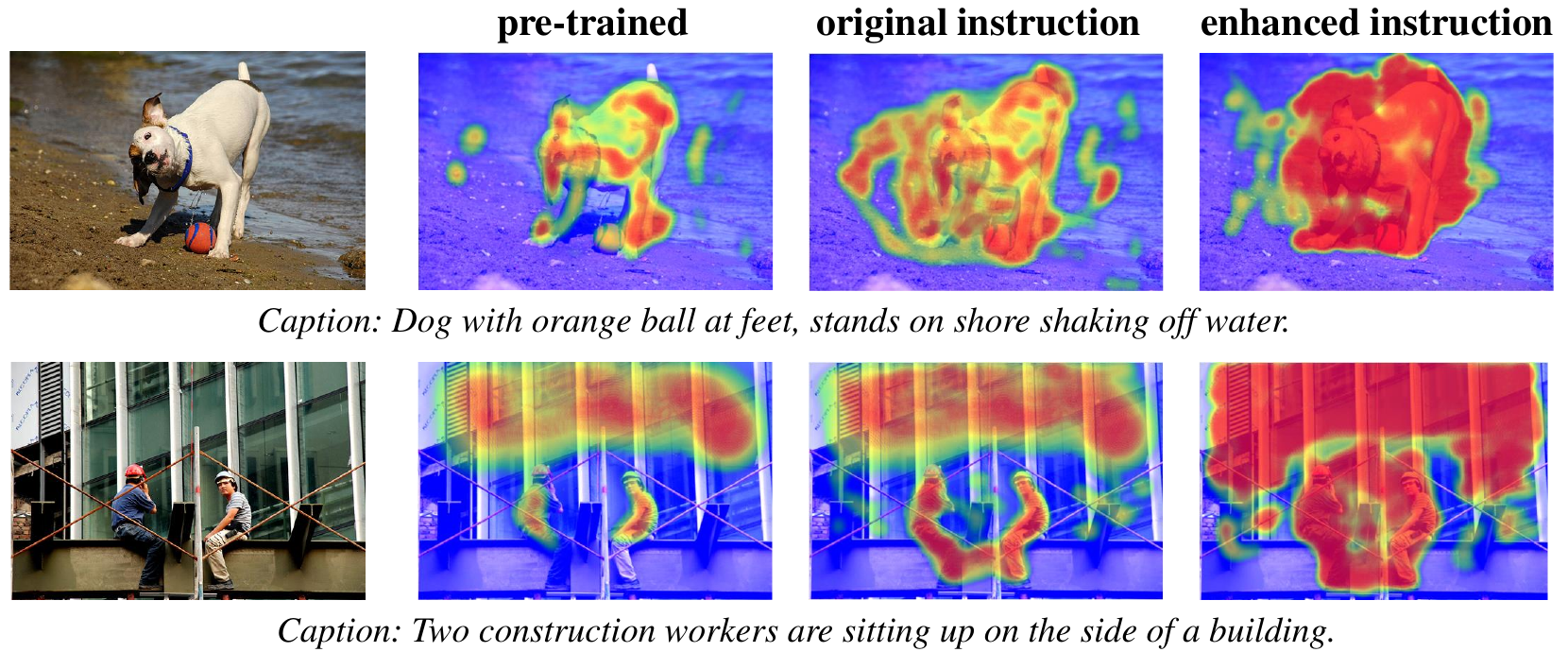}
    \vspace{-0.3cm}
    \caption{Cross-attention visualization results on Flickr30k dataset.}
    \label{fig:crossattention} 
    \vspace{-0.5cm}
\end{figure*}

\subsection{Ablation Study}
This section aims to verify the effectiveness of the important components with comprehensive ablation studies. We introduce five model variants as follows:
(1) \textbf{PETAL} V1 removes the adaptive instruction MOEs and feeds the original template directly.
(2) \textbf{PETAL} V2 removes the score-based IB loss in Eq. \ref{eq:IB} for guiding the learning.
(3) \textbf{PETAL} V3 removes both the adaptive instruction MOEs and the score-based IB loss in Eq. \ref{eq:IB}.
(4) \textbf{PETAL} V4 removes the dynamic threshold $\Gamma$ in Eq. \ref{eq:dma}.
(5) \textbf{PETAL} w. random instruction just utilizes the random learnable vector as the enhanced instruction.
Additionally, we retain the original version of \method{} as a control. These experiments are conducted on the AOKVQA and Flickr30K datasets, as illustrated in Table \ref{tab:ablation}. 
According to the results above, we can draw some conclusions as follows:
\begin{itemize}[leftmargin=*]
    \item Experimental results reveal a significant drop of 1.1 in overall performance when the adaptive instruction MOEs module is removed. This highlights the importance of finer-grained instructions in extracting more precise image features. The presence of the information mixture module strengthens the interoperability between images and text, which is crucial for the successful completion of multimodal tasks.
    \item Excluding the score-based IB loss results in a marginal reduction with an average decrease of 0.4, emphasizing the effective role of mutual information in multimodal tasks. The information bottleneck module aids in capturing correlated information between different modalities.
    \item Removing both the adaptive instruction MOEs and the score-based IB loss leads to a more significant performance decline, averaging a 3.0 decrease. This indicates that these components collectively contribute to the model's robustness.
    \item Removing the dynamic threshold $\Gamma$, this configuration experiences an average performance fall of 1.3, potentially leading to a less nuanced application of the pre-trained weights and less effective learning outcomes.
    \item Utilizing a random learnable vector as the enhanced instruction results in an average performance decrease of 2.2. This suggests that the content of the instruction itself plays a significant role and cannot be simply replaced by introducing random noise, which highlights the value of well-structured instruction processing.
\end{itemize}

\subsection{Visualization Analysis}

To gain a more intuitive understanding of instruction enhancement principles, we conduct a visualization analysis of the interaction process within the parameter-efficient tuning process. Specifically, we examine the attention scores assigned to different image patches by the instructions and represent these scores using different color shades on the image. As depicted in Figure~\ref{fig:crossattention}, different instructions exhibit varying attention points for image features. Furthermore, a comparison with the pre-enhanced instructions reveals that enhanced instructions more accurately capture essential image features. This enhanced feature representation is highly beneficial for subsequent tasks such as question answering and other multimodal tasks.

\begin{table}[h]
\caption{Training time and parameter size comparison.}
\label{tab:cost}
\centering
\scriptsize
\vspace{-0.3cm}
\setlength{\tabcolsep}{1.2mm}{
\begin{tabular}{lc|ccccc}
\toprule[1pt]
Method & \#Tunable & Flickr30K & TextCaps & AOKVQA & OKVQA & TextVQA \\ \midrule
Fine-tune & 188M & 1.00 & 1.00 & 1.00 & 1.00 & 1.00 \\
MAPLE & 4.8M & 0.95 & 0.93 & 0.94 & 0.90 & 0.91 \\
LoRA & 5.0M & 0.91 & 0.92 & 0.86 & 0.87 & 0.93 \\
PETAL & 1M & 0.85 & 0.88 & 0.86 & 0.85 & 0.91 \\ \bottomrule[1pt]
\end{tabular}}
\vspace{-0.5cm}
\end{table}

\subsection{Time Cost}

The results in Table \ref{tab:cost} provides a comprehensive comparison of training times and parameter sizes for image captioning and VQA tasks. In this table, the baseline for both training time and parameter cost is set by the full fine-tune InstructBLIP , which is normalized to a unit value of 1.00 for each of the datasets.

A key observation from the table is the significantly reduced number of tunable parameters and faster training times achieved by our PETAL. PETAL, with only 1M tunable parameters, demonstrates an impressive efficiency, achieving training times ranging from 0.85 to 0.91 times the baseline across the datasets. This is substantially lower than the full fine-tune method.

\section{Conclusion}
\label{sec:conclusion}

This paper presents PETAL, a novel approach for parameter-efficient tuning in vision-language models. We have demonstrated that PETAL, with its dynamic mode approximation and enhanced instruction semantics through an adaptive instruction MOEs and a mutual information loss, offers a significant leap forward. Our experiments on various tasks validate PETAL's superior performance over existing methods on benchmark datasets, achieving notable improvements with minimal computational resources. PETAL marks a critical step in making advanced vision-language models more accessible and practical. Future work could explore further optimizations in the framework and extend its application to other domains, potentially revolutionizing how we approach and utilize large multimodal models in various fields.

\clearpage


\clearpage

\maketitlesupplementary

\appendix

\begin{table*}[!ht]
\caption{A comprehensive review of the datasets.}
\centering
\scriptsize
\label{tab:efficient_finetune_methods}
\begin{tabular}{@{}ccc@{}}

\toprule[1pt]
\textbf{Image Sample} & \textbf{Data Template} & \textbf{Demonstration} \\ \midrule
\multicolumn{3}{c}{\textbf{Datasets for Visual Question Answering}} \\\midrule
   \begin{minipage}{.19\textwidth}
      \includegraphics[width=\linewidth]{figure/okvqa.png}
      
      \quad\qquad\qquad \textbf{OKVQA}
    \end{minipage} 
& 
\begin{minipage}{.33\textwidth}
\textcolor{cyan}{\textbf{Instruction:}} \\
\textbf{What fruit is typically added to the top of cereal?} \\
\\

\textcolor{cyan}{\textbf{Answer:}} \\ 
\textbf{``banana", ``banana", ``banana", ``banana", ``banana", ``banana", ``strawberry", ``strawberry", ``blueberry", ``blueberry"}
    \vspace{3pt}
    \end{minipage} 
& 
 \begin{minipage}{.45\textwidth}
    OK-VQA stands as a dataset designed specifically for visual question answering, challenging methods to leverage external information for accurate responses. It comprises 14,055 open-ended questions, each associated with five ground truth answers. To ensure the necessity of external knowledge, the dataset underwent meticulous manual curation, filtering questions to mandate reliance on external sources such as Wikipedia. Additionally, the dataset has been refined by reducing questions that commonly elicit the same answers, thereby mitigating potential biases within the dataset.
   \end{minipage}  
\\ \midrule

   \begin{minipage}{.19\textwidth}
      \includegraphics[width=\linewidth]{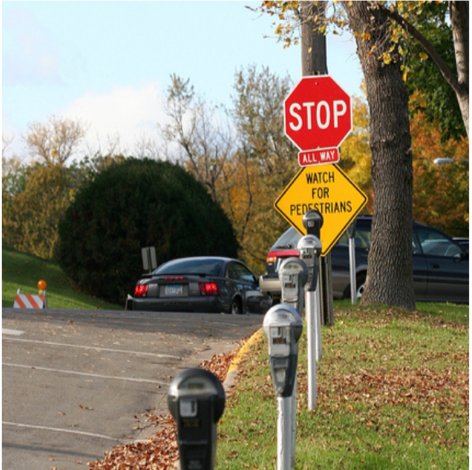}
      
      \quad\qquad\qquad \textbf{A-OKVQA}
    \end{minipage} 
& 
\begin{minipage}{.33\textwidth}
\textcolor{cyan}{\textbf{Instruction:}} \\
\textbf{What does the yellow sign advise you to watch for?} \\
\\

\textcolor{cyan}{\textbf{Answer:}} \\ 
\textbf{``parking", ``watch", ``pedestrians", ``people", ``pedestrian", ``pedestrians", ``pedestrians", ``signal board", ``pedestrians", ``warning"}
\vspace{3pt}
\end{minipage} 
    
& 
 \begin{minipage}{.45\textwidth}
    A-OKVQA is a knowledge-based visual question answering benchmark. A-OKVQA is an Augmented successor of OK-VQA and contains a diverse set of 25K questions requiring a broad base of commonsense and world knowledge to answer. Questions in A-OKVQA are challenging, conceptually diverse, require knowledge outside the image, and in contrast to existing knowledge-based visual question answering datasets, they cannot be answered by simply querying a knowledge base. To ease working with unbounded knowledge sources, questions in the training set are paired with rationales that supply facts and snippets of reasoning needed to answer them.
   \end{minipage}  
\\ \midrule
   \begin{minipage}{.19\textwidth}
      \includegraphics[width=\linewidth]{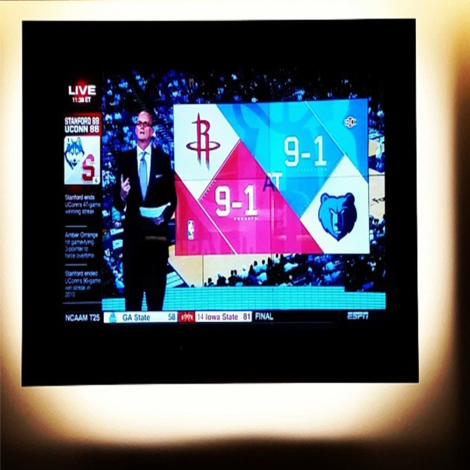}
      
      \quad\qquad\qquad \textbf{TextVQA}
    \end{minipage} 
& 
\begin{minipage}{.33\textwidth}
\textcolor{cyan}{\textbf{Instruction:}} \\
\textbf{When is this being aired?} \\
\\

\textcolor{cyan}{\textbf{Answer:}} \\ 
\textbf{``11:38 et", ``unanswerable", ``unanswerable", ``live", ``11:38 et", ``unanswerable", ``live 11:39 eastern time", ``live", ``live", ``no text in image"}
\vspace{3pt}
\end{minipage} 
    
& 
 \begin{minipage}{.45\textwidth}
    TextVQA challenges models to not only interpret but also reason about textual content within images to accurately respond to related questions. This task demands that models integrate a novel modality—the text embedded in images—and apply logical reasoning over this text to successfully address TextVQA queries. Comprising a robust dataset of 28,408 images sourced from OpenImages, TextVQA is accompanied by a diverse set of 45,336 questions. These questions are designed to probe a wide range of textual understanding and reasoning skills. Each question comes with multiple valid answers, cumulatively amounting to an impressive total of 453,360 ground truth responses. 
   \end{minipage}  
\\ \midrule
\multicolumn{3}{c}{\textbf{Datasets for Image Captioning}} \\\midrule
   \begin{minipage}{.19\textwidth}
      \includegraphics[width=\linewidth]{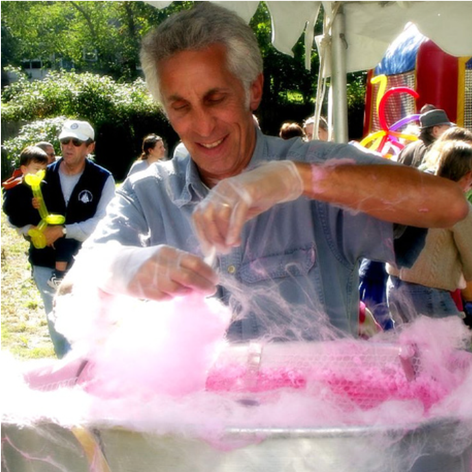}
      
      \quad\qquad\qquad \textbf{Flickr30K}
    \end{minipage} 
& 
\begin{minipage}{.33\textwidth}
\textcolor{cyan}{\textbf{Instruction:}} \\
\textbf{Describe this picture} \\
\\

\textcolor{cyan}{\textbf{Answer:}} \\ 
\textbf{``A tanned man with white hair wearing plastic gloves making cotton candy in front of a small fair with parents and children.”, ``An older man is making cotton candy while others look on.”, ``At some sort of carnival, a man is making cotton candy.”, ``A man in a blue shirt making pink cotton candy.”}
\vspace{3pt}
\end{minipage} 
    
& 
 \begin{minipage}{.45\textwidth}
    The Flickr30k dataset encompasses a collection of 31,000 images meticulously curated from Flickr, each complemented by five descriptive reference sentences meticulously crafted by human annotators. This dataset has established itself as a benchmark standard in the realm of sentence-based image description, providing a rich resource for evaluating and enhancing natural language processing and computer vision models. Its diverse range of images paired with multiple human-authored sentences per image offers a comprehensive platform for training and testing sophisticated models in image understanding and language generation. The inclusion of human-generated descriptions not only adds a layer of complexity but also injects a valuable element of human perception and linguistic variation into the mix.
   \end{minipage}  
\\ \midrule

   \begin{minipage}{.19\textwidth}
      \includegraphics[width=\linewidth]{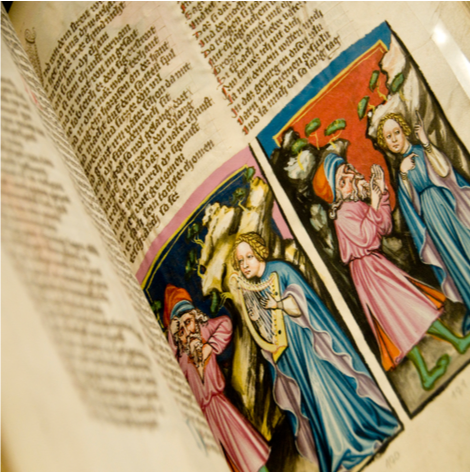}
      
      \quad\qquad\qquad \textbf{TextCaps}
    \end{minipage} 
& 
\begin{minipage}{.33\textwidth}
\textcolor{cyan}{\textbf{Instruction:}} \\
\textbf{Describe this picture.} \\
\\

\textcolor{cyan}{\textbf{Answer:}} \\ 
\textbf{``some strange writing is next to a painting.”, ``A book is written in German and has colorful images of people from mythological times playing a harp.”, ``An old book is open to a page that ends with the word tag.”, ``An ancient book has a page that ends with the word tag.”, ``A book features pictures and text written in a non English language”}
\vspace{3pt}
\end{minipage} 
    
& 
 \begin{minipage}{.45\textwidth}
    TextCaps presents a unique challenge for models, requiring them to not only read but also interpret and reason about textual information embedded within images to generate meaningful captions. This task necessitates the integration of a novel textual modality found in the images, demanding that models simultaneously process and synthesize both textual and visual content to create coherent and contextually relevant image descriptions. Leveraging a comprehensive dataset of 28,408 images sourced from OpenImages, TextCaps offers a rich trove of 142,040 captions, averaging five distinct captions per image. This extensive collection of images and captions provides a robust testing ground for models to refine and demonstrate their capabilities in understanding and contextualizing text within a visual framework.

   \end{minipage}  
\\
\bottomrule[1pt]
\end{tabular}

\end{table*}

\section{Source Code}
Our source code is available at: \url{https://github.com/melonking32/PETAL}.

\section{Experimental Details}
\subsection{Details of Baselines}
\noindent\textbf{InstructBLIP}~\cite{dai2023instructblip} is an extension on BLIP2 that further improves performance through instruction fine-tuning. InstructBLIP consists of a visual encoder, a LLM, and a cross-modal bridging Q-former. During training, the visual encoder and LLM are frozen and only the Q-former parameters are updated to adapt to downstream tasks. This mode can connect multiple pre-trained visual models and LLMs. We use the parameters of InstructBLIP to initialize the model, fine-tuning it on downstream tasks.

\noindent\textbf{LLaVA}~\cite{liu2023improvedllava}
builds 158K multimodal instruction-tuning data using GPT-4 and utilizes this dataset to train a novel multi-modal model. Architecturally, visual embeddings are transformed into text embeddings through a simple linear layer. These embeddings are then concatenated with instructions and fed into the LLaMA Decoder for end-to-end training. LLaVA achieves state-of-the-art performance on the Science QA dataset while demonstrating robust and versatile multi-modal interaction capabilities.

\noindent\textbf{Head Tuning}~\cite{zaken2021bitfit} posits that when applying pre-trained models to downstream tasks, updating only the head parameters suffices. In our experiment, we maintain the other layer parameters as fixed and exclusively update the parameters of the final transformer layer.

\noindent\textbf{LLaMA-Adapter}~\cite{gao2023llama}
addresses the efficiency concerns of multi-modal models by integrating the adapter into the LLaMA architecture and deploying it in the multi-modal domain. Moreover, to enhance its performance in open-ended visual instruction scenarios, LLaMA-Adapter introduces more learnable parameters (e.g., norm, bias, and scale) and implements an early fusion strategy for better integration of visual knowledge. It also adapts a joint training paradigm for image-text pairs and instruction-following. LLaMA-Adapter demonstrates enhanced performance in open-ended multi-modal instructions, requiring only a modest increase in parameters compared to LLaMA. Additionally, it displays stronger language-only instruction-following abilities and excels in chat interactions.

\noindent\textbf{MAPLE}~\cite{khattak2022maple} aims to address the sensitivity of multi-modal tasks to prompt templates and the insufficient interaction across modalities. It proposes the use of learnable prompts to enhance alignment between vision and language representations. The trained model exhibits promising performance and generalization on downstream tasks. In our experiments, following the same approach, we incorporate learnable prompts at each layer.

\noindent\textbf{LoRA}~\cite{hu2021lora} is added in parallel to the original parameter matrix. These parameters result from rank decomposition of the original parameter matrix. Similarly, the representation is first reduced to a lower dimension, then elevated back to the original dimension through dimensionality expansion. The output from the main network and Lora are summed and propagated backward together. During training, only the parameters of LoRA are updated. We employ Kaiming Initialization \cite{He2015DelvingDI} to initialize the weights of the dimension reduction layer and use zero initialization for bias as well as the weights of the dimension expansion layer. Specifically, we set the rank of LoRA to 64 for the experiment.

\subsection{Details of Datasets}
Please refer to Table \ref{tab:efficient_finetune_methods}
for details. 

\begin{table*}[t]
\caption{Instruction templates we use, instructions are employed to extract richer information from images, while task-specific instructions are utilized to guide the LLM in completing downstream tasks.}
\centering
\label{tab:instrct}
\scriptsize
\setlength{\tabcolsep}{6.5mm}{
\begin{tabular}{l|l|l}
\toprule[1pt]
Task & Type & Instruction Template \\ \midrule
\multirow{4}{*}{\begin{tabular}[c]{@{}l@{}}Image Captioning\end{tabular}} & \multirow{3}{*}{\begin{tabular}[c]{@{}l@{}} Original Instruction  \end{tabular}} & What objects are in the picture? \\
 &  & What are the characteristics of the objects in the picture \\
 &  & What is the relationship between the objects in the picture? \\ \cmidrule{2-3} 
 & \begin{tabular}[c]{@{}l@{}} Task Specific Instruction\end{tabular} & A photo of \{ \} \\ \midrule
\multirow{4}{*}{VQA} & \multirow{3}{*}{\begin{tabular}[c]{@{}l@{}} Original Instruction \end{tabular}} & What objects are in the picture? \\
 &  & What color are the objects in the picture? \\
 &  & What are the characteristics of the objects in the picture \\ \cmidrule{2-3} 
 & \begin{tabular}[c]{@{}l@{}}Task Specific Instruction\end{tabular} & Question: \{Question\} Short answer: \\ \bottomrule[1pt]
\end{tabular}}
\end{table*}

\begin{figure*}[t]
  \centering
  \includegraphics[width=\textwidth]{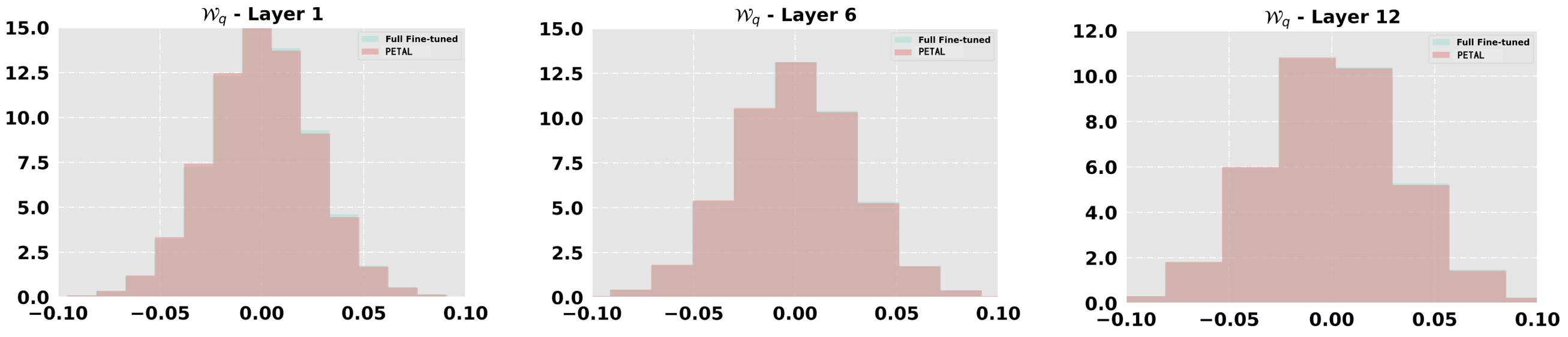}
  \caption{Parameter Distribution. From left to right are the comparisons of parameter distributions after training with PETAL and after Full Fine-tune for the 1st layer, 6th layer, and 12th layer.}
  \label{fig:para}
\end{figure*}

\begin{figure*}[t]
  \centering
  \includegraphics[width=1\textwidth]{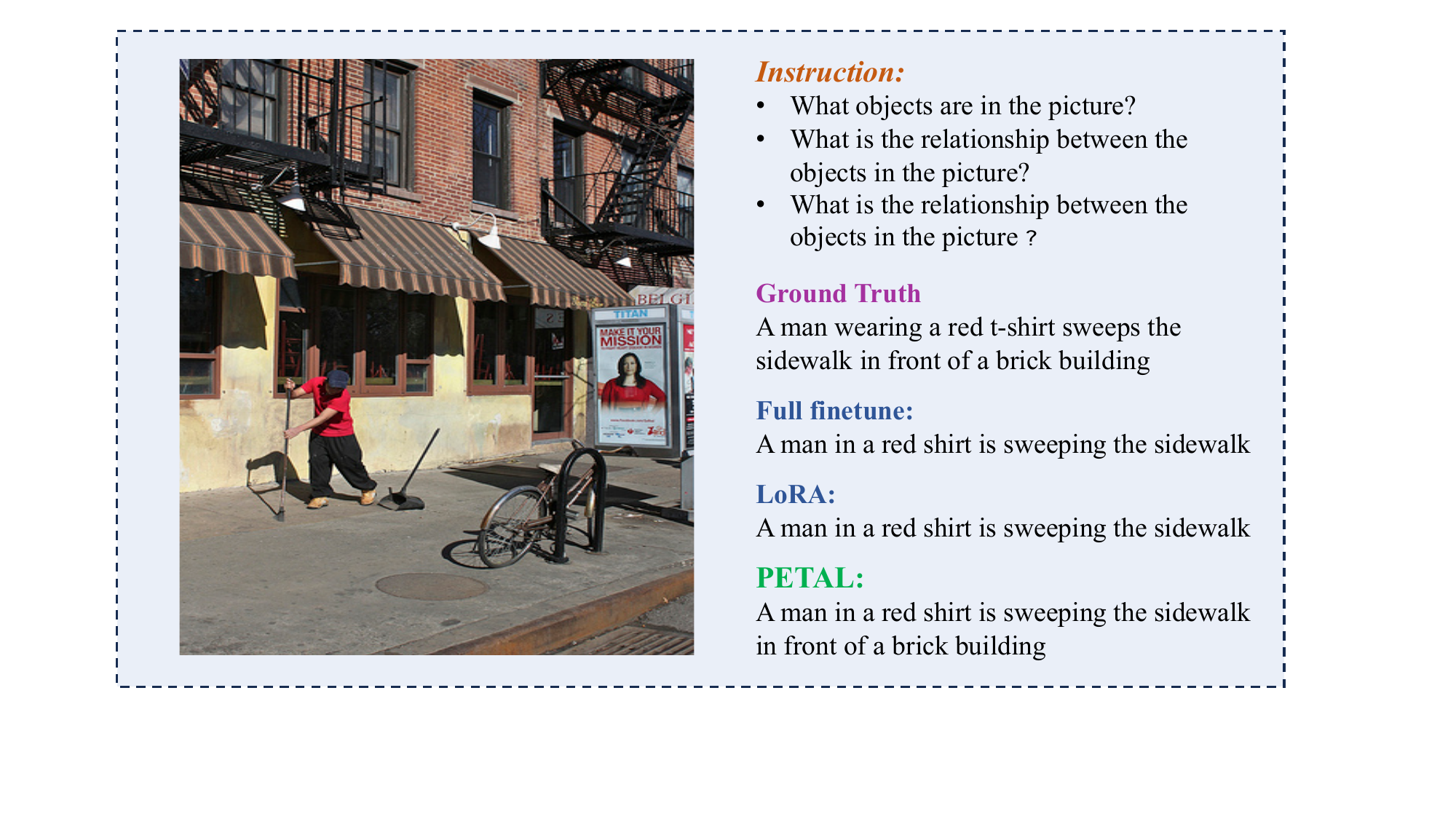}
  \caption{It is obvious that PETAL can capture more object information than other methods, thus providing a more comprehensive understanding of the image.}
  \label{fig:case1}
\end{figure*}

\subsection{Evaluation Metric}
\noindent\textbf{Image Captioning} is the task of describing the content of an image in words. Each image has a certain number of reference captions. We select several commonly used metrics in captioning task to measure the gap between the caption generated by the model and the candidate captions.

\begin{itemize}[leftmargin=*]
\item  \textbf {CIDEr}~\cite{Vedantam2014CIDErCI}  (Consensus-based Image Description Evaluation) metric is a way of evaluating the quality of generated textual descriptions of images. The CIDEr metric measures the similarity between a generated caption and the reference captions, and it is based on the concept of consensus: the idea that good captions should not only be similar to the reference captions in terms of word choice and grammar, but also in terms of meaning and content. The generated caption is compared to each reference caption using the BLEU (Bilingual Evaluation Understudy) score, which measures the n-gram overlap between the generated caption and the reference captions. The BLEU scores are then modified using an IDF (Inverse Document Frequency) weighting, which gives more weight to words that are rare in the reference captions but appear in the generated caption. Finally, the weighted BLEU scores are averaged over all reference captions to produce the final CIDEr score. 
\item  \textbf {ROUGE}~\cite{Lin2004ROUGEAP} (Recall-Oriented Understudy for Gisting Evaluation) is also a metric used to measure the correlation between the generated caption and the reference captions. It has many branches, in our experiments, we chose \textbf{ROUGE-1} to calculate the overlap of uni-grams (each word) between the generated caption and reference captions. Additionally, \textbf{ROUGE-L} naturally considers sentence-level structure similarity and automatically identifies the longest co-occurring in-sequence n-grams. We compute the Recall and F1 score for each metric.
    
\end{itemize}

\noindent\textbf{VQA}(Visual Question Answering) is a task in computer vision that entails answering questions about an image. The objective of VQA is to instruct machines to comprehend the content of an image and respond to questions about it using natural language. For our VQA dataset, each question has 10 candidate answers. During evaluation, if the model's predicted answer matches any 3 of them exactly, we consider the answer as correct. The final calculation involves accumulating the accuracy (ACC) metric.


\section{Instruction Details}
In Table \ref{tab:instrct}, we present the specific instruction templates employed in the training of our model, designed to extract richer information from images and guide the LLM in completing various downstream tasks. These templates are categorized based on the task type—namely Image Captioning and VQA.
For Image Captioning, the original instruction prompts are crafted to elicit detailed descriptions of the images, encouraging the model to recognize and articulate not just the objects present, but also their attributes and the interactions between them. The `Task Specific Instruction' for Image Captioning simplifies this to a more direct format, "A photo of \{\}", allowing for more focused responses.

In the VQA category, similar `Original Instruction' templates are used, such as ``What color are the objects in the picture?”, alongside the aforementioned one. These are designed to prompt the model to focus on specific details within the image, thus aiding in developing a more nuanced understanding of the visual content. The `Task Specific Instruction' for VQA takes the form of ``Question: \{Question\} Short answer:”, tailoring the model's response to concise, direct answers to specific questions posed about the image.
This structured approach to instruction design not only enhances the model's capability in interpreting visual data but also fine-tunes its responsiveness to specific types of queries, thereby broadening its applicability across different tasks and scenarios in image understanding and language processing.

\section{Parameter Distribution Visualization}
In Figure \ref{fig:para}, we compare the model parameters after training with PETAL and after full fine-tune. PETAL progressively incorporates knowledge learned from downstream tasks into the parameters of the pre-trained model during training, making minor adjustments to the original parameters. It can be observed that in the initial layers, there is a small difference in parameter distributions between PETAL and full fine-tune. However, in the later layers, the parameter distributions of PETAL closely align with those of full fine-tune. Notably, we require fewer trainable parameters, and the slight differences in parameters lead to performance improvements on the new task.

\begin{figure*}[t]
  \centering
  \includegraphics[width=\textwidth]{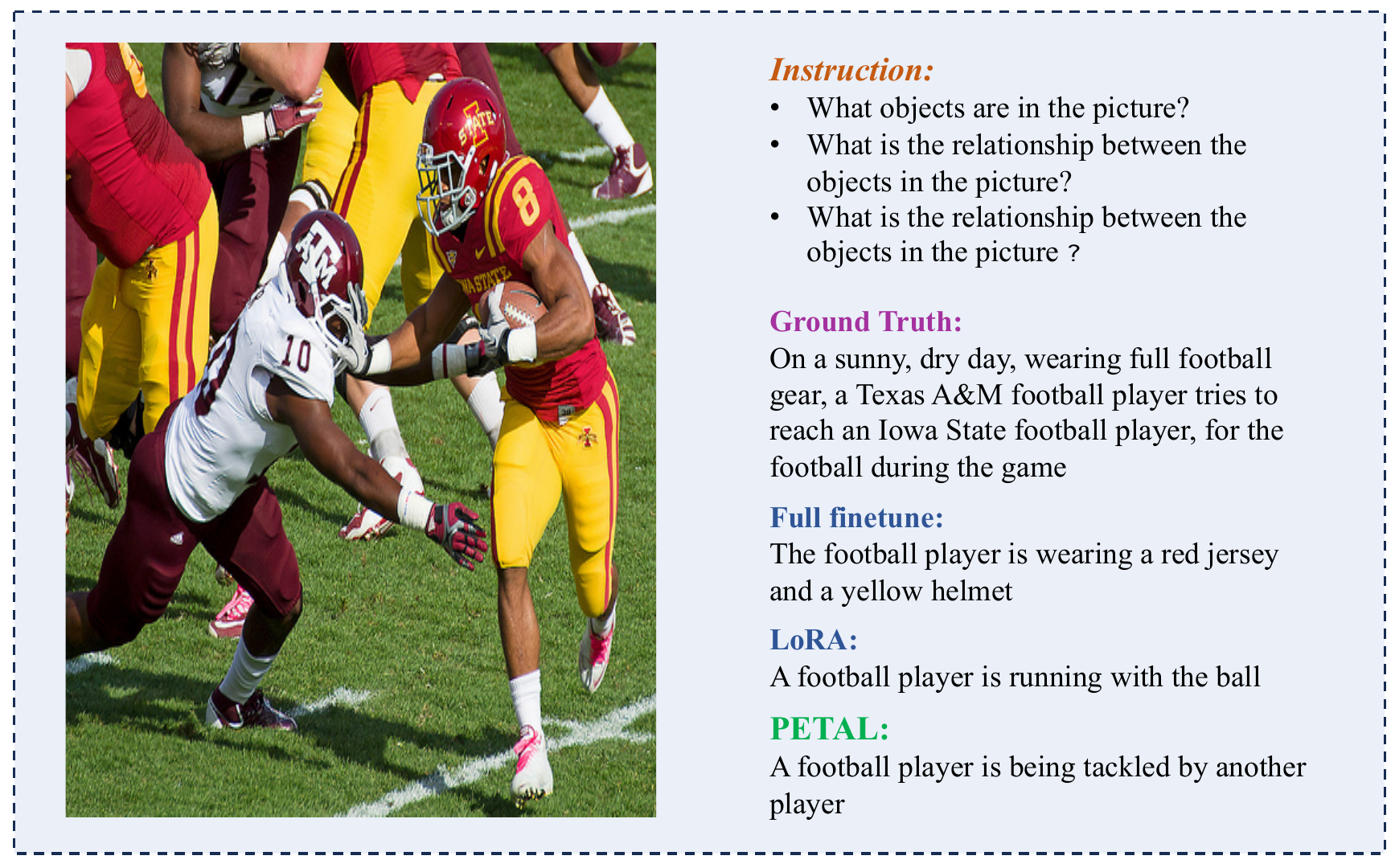}
  \caption{Under the guidance of our enhanced instruction, PETAL can not only accurately capture the features of object in the image, but also analyze the relationships between objects.}
  \label{fig:case4}
\end{figure*}

\section{Details of Parameter Size}
In the pursuit of optimizing model performance with minimal parameter overhead, our parameter-efficient instruction tuning PETAL marks a significant stride. By strategically tuning a fraction of the total parameters, we achieve comparable or superior results. Specifically, our PETAL requires only about 1/188 of the parameter count used by the Q-Former in InstructBLIP.

Let the hidden size of the image feature be denoted as $\mathbf{H}_v$, the text feature as $\mathbf{H}_t$, the rank setting as $\mathbf{R}$, and the middle size of an expert network as $\mathbf{M}$. The trainable parameters can be categorized into two distinct parts: the Q-former and the MOEs. The parameter count for the Q-former part is calculated as:
\begin{equation}
    \begin{split}
        &\left( \mathbf{H}_v \times \mathbf{R} + \mathbf{H}_t \times \mathbf{R} + \mathbf{R} \times \mathbf{R} \times \mathbf{R} \right) \\
        +& \left( \mathbf{H}_t \times \mathbf{R} + \mathbf{H}_t \times \mathbf{R} + \mathbf{R} \times \mathbf{R} \times \mathbf{R} \right)\\
        =& \left( 
        \mathbf{H}_v + 2 \mathbf{R}^2 +3 \mathbf{H}_t \right) \times \mathbf{R}
    \end{split}
\end{equation}
In this formulation, the first row represents the entirety of the shared approximation matrix within the cross-attention module. Specifically, the initial term corresponds to the factor matrix $\mathbf{U}$, as detailed in Section \ref{sec:dma}, while the subsequent two terms are associated respectively with $\mathbf{V}$ and $\mathbf{P}$. The second row mirrors this structure, applying the same operational sequence to the self-attention module. 

And the parameter size for the MOEs is: $\mathbf{H}_t \times \mathbf{M}  \times 2 \times 3$, where 2 is for the encode-decode architecture with 2 layers, and 3 is the number of experts. In our experiments, with $\mathbf{H}_v = 1408$, $\mathbf{H}_t = 768$, $\mathbf{R} = 64$, and $\mathbf{M} = 64$, the total number of parameters sums up to $1408 \times 64 + 768 \times 64 \times 3 + 64 \times 64 \times 64 \times 2 + 768 \times 64 \times 2 \times 3 = 1056768$, which is approximately 1M.

\section{Case Study}
We visualize the case studies of our PETAL and the baselines in Figure \ref{fig:case1} and Figure \ref{fig:case4}. Based on the results from the two case studies, the PETAL model demonstrates several advantages in processing and interpreting images for text generation tasks. Specifically, in the first image, where a man in a red shirt is sweeping the sidewalk, PETAL not only identifies the action and the subject but also correctly associates the activity with the brick building in the background. This indicates that PETAL is adept at contextually analyzing the scene and incorporating relevant environmental details into its descriptions, surpassing the simpler descriptions generated by Full fine-tune and LoRA. In the second case, PETAL's description captures the dynamic action in the image by recognizing the interaction between the football players, specifically the tackling event. This suggests that PETAL has an enhanced ability to detect and describe interactions and motions within images, which is a complex aspect of image understanding that can be challenging for models. Besides, PETAL's descriptions align closely with the provided ground truth for both images. This implies a high degree of accuracy in PETAL's language generation capabilities, suggesting that it can be reliably used in applications where precise and accurate image descriptions are crucial.

\clearpage
{
    \small
    \bibliographystyle{ieeenat_fullname}
    \bibliography{main}
}

\end{document}